\documentclass{article}

\usepackage[nonatbib, preprint]{neurips_2024}

\usepackage[utf8]{inputenc} % allow utf-8 input
\usepackage[T1]{fontenc}    % use 8-bit T1 fonts
\usepackage{hyperref}       % hyperlinks
\usepackage{url}            % simple URL typesetting
\usepackage{booktabs}       % professional-quality tables
\usepackage{amsfonts}       % blackboard math symbols
\usepackage{nicefrac}       % compact symbols for 1/2, etc.
\usepackage{microtype}      % microtypography
\usepackage{xcolor}         % colors
\usepackage{graphicx}
\usepackage{subfig}
\usepackage{colortbl}  %彩色表格需要加载的宏包
\usepackage{xcolor}
\usepackage{booktabs,multirow,makecell,array}
\usepackage{makecell}
\usepackage{algorithm}  
\usepackage{algorithmicx}
\usepackage{algpseudocode}  
\usepackage[figuresleft]{rotating}
\usepackage{wrapfig}
\usepackage{CJKutf8}
\usepackage{enumitem}
\title{Pay More Attention to the Robustness of Prompt for Instruction Data Mining}

\author{%
  Qiang Wang, \hspace{1mm} Dawei Feng, \hspace{1mm}Xu Zhang, \hspace{1mm}Ao Shen, \hspace{1mm}Yang Xu, \hspace{1mm}Bo Ding, \hspace{1mm}Huaimin Wang \vspace{2mm}\\
  \vspace{2mm}
  National University of Defense Technology\\
  \vspace{2mm}
  Hunan Changsha 410073, China \\
  \vspace{2mm}
  \tt \href{mailto:happytiger_95@163.com}{happytiger\_95@163.com} \\
}

\begin{document}
\begin{CJK}{UTF8}{gbsn}

\maketitle

\begin{abstract}
 Instruction tuning has emerged as a paramount method for tailoring the behaviors of LLMs. Recent work has unveiled the potential for LLMs to achieve high performance through fine-tuning with a limited quantity of high-quality instruction data. Building upon this approach, we further explore the impact of prompt's robustness on the selection of high-quality instruction data. This paper proposes a pioneering framework of high-quality online instruction data mining for instruction tuning, focusing on the impact of prompt's robustness on the data mining process. Our notable innovation, is to generate the adversarial instruction data by conducting the attack for the prompt of online instruction data. Then, we introduce an Adversarial Instruction-Following Difficulty metric to measure how much help the adversarial instruction data can provide to the generation of the corresponding response. Apart from it, we propose a novel Adversarial Instruction Output Embedding Consistency approach to select high-quality online instruction data. We conduct extensive experiments on two benchmark datasets to assess the performance. The experimental results serve to underscore the effectiveness of our proposed two methods. Moreover, the results underscore the critical practical significance of considering prompt‘s robustness.
\end{abstract}

\section{Introduction}

The emergence and application of Large Language Models (LLMs)~\cite{radford2018improving, radford2019language,NEURIPS2020_1457c0d6,achiam2023gpt} have brought about revolutionary changes in the field of artificial intelligence, across diverse domains such as code completion~\cite{liu2020multi}, question answering~\cite{yan2021large}, summery generation~\cite{dong2019unified}, language translation~\cite{chen2024general2specialized}, and conversational agents~\cite{deng2023rethinking}. However, LLMs often encounter difficulties in understanding human intent and cannot generate appropriate responses~\cite{zhang2023instruction, bakker2022fine}. To tackle this challenge, instruction tuning has emerged as a paramount method for customizing the behaviors of LLMs. The core of Instruction tuning~\cite{shu2023exploitability,longpre2023flan,peng2023instruction,ouyang2022training} is to leverage meticulously crafted instruction data to direct LLMs in generating responses that are in line with human expectations.

However, the construction of instruction data usually requires experts to manually calibrate, and massive amounts of data requires high cost and resources. Therefore, instruction data mining has emerged as a promising approach for selecting high-quality data to enhance the LLM's performance. Recent studies found that LLMs can be fine-tuned to perform well even with a small amount of high-quality instruction data~\cite{zhou2024lima,du2023mods}. The most representative method for instruction tuning is based on self-guided, denoting the Instruction-Following Difficulty (IFD) as the metric to select the high-quality instruction data for instruction tuning ~\cite{li2023quantity}. 
Taking inspiration from this approach, we further consider the impact of prompt's robustness on the selection of high-quality instruction data.

The prompt serves as an intermediary mechanism linking human input with LLMs, can express tasks that are less clearly defined in natural language in a way that LLMs can understand, so the quality of prompt directly affects the performance of LLMs.
However, LLMs are known to be highly sensitive to prompt, e.g, the order of few-shot examples, minor typos, or different expressions with the same semantic meaning can lead to entirely different results~\cite{lu-etal-2022-fantastically,maus2023black,SiGYWWBW23,0001HH0ZWY0HGJ024}. Therefore, it is necessary and urgent to analyze the impact of prompt' robustness on instruction data mining.

\begin{figure*}[ht]
\label{framework_scene}
\centering
\includegraphics[scale=0.46]{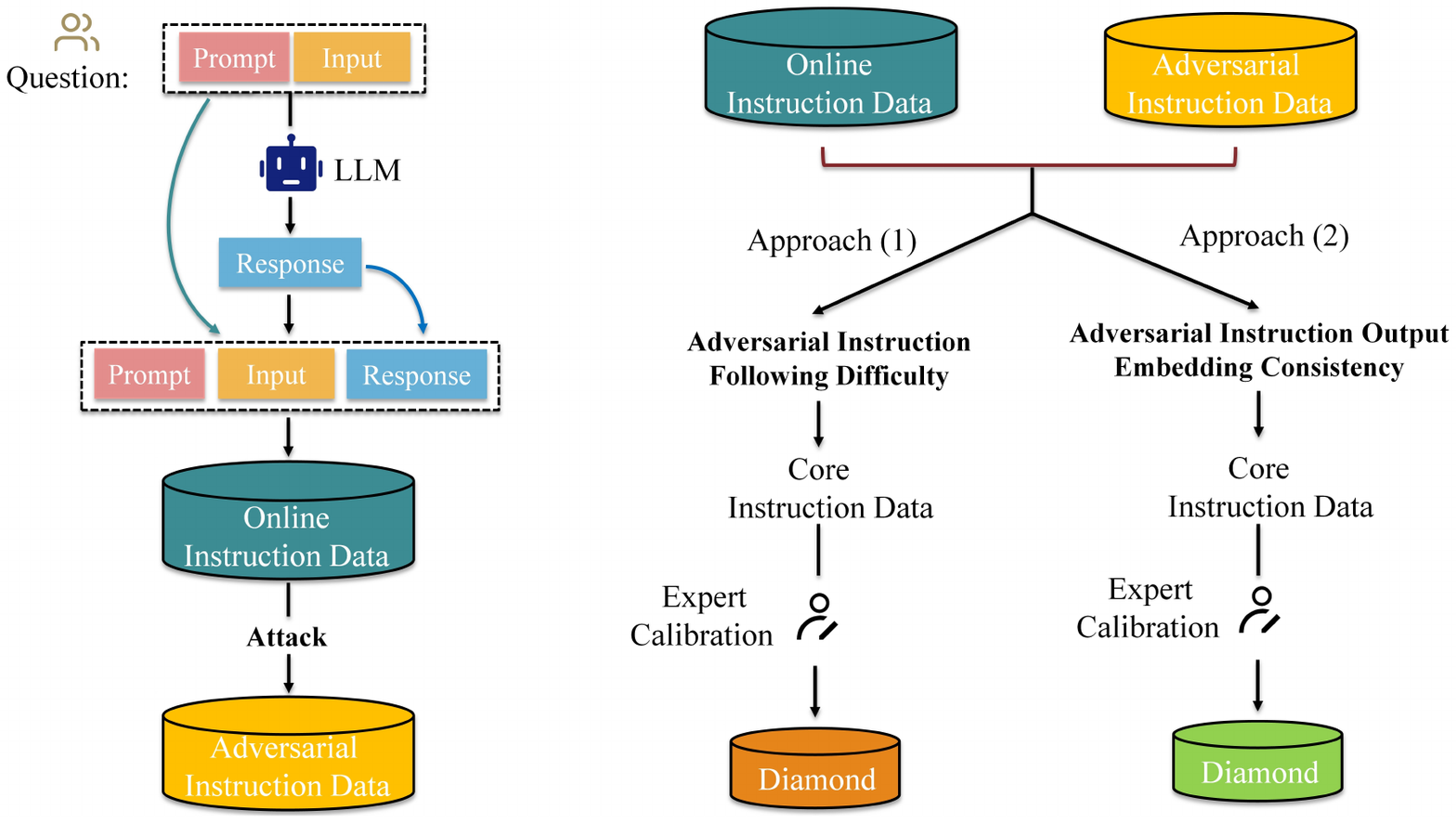}
\caption{\label{framework} The illustration of the overall framework of diamond samples mining from online instruction data to conduct the instruction tuning.}
\label{fig:framework}
\end{figure*}

% Taking inspiration from the direct correlation observed between the robustness of online user input prompts and the quality of responses, we explore the influence of prompt's robustness on the selection of high-quality samples.

Figure ~\ref{fig:framework} illustrates the overall framework of 
high-quality instruction (named as diamond ) data mining from online instruction data for the instruction tuning, integrating two novel self-guided approaches to select diamond data. We perform three different types of attacks, the character-level~\cite{LiJDLW19,gao2018black}, word-level~\cite{JinJZS20,LiMGXQ20}, and sentence-level~\cite{NaikRSRN18,RibeiroWG021} for the user's prompt, to generate different adversarial instruction samples~\cite{zhu2023promptbench}. Based on the adversarial instruction samples and IFD algorithm, we propose a novel Adversarial Instruction-Following Difficulty (AIFD) score, a self-guided approach enabling models to autonomously select diamond data from online instruction data. The higher AIFD score of instruction data, indicating less guidance and lower robustness, the more it should be mined for instruction tuning. Then, we introduce a novel Adversarial Instruction Output Embedding Consistency as the metric to mine the diamond data, through measuring the output embedding similarity between the adversarial prompt and user input' prompt.

In this paper, the main contributions of our work are summarized as follows:
\begin{itemize}[leftmargin=*]
%     \item You should answer \answerYes{}, \answerNo{}, or \answerNA{}.
%     \item \answerNA{} means either that the question is Not Applicable for that particular paper or the relevant information is Not Available.
%     \item Please provide a short (1–2 sentence) justification right after your answer (even for NA). 
%    % \item {\bf The papers not including the checklist will be desk rejected.}
% ({\romannumeral1}) 
\item
We propose a novel framework for diamond data mining from online instruction data for instruction tuning. Considering the influence of prompt's robustness for instruction data, we introduce the Adversarial Instruction-Following Difficulty (AIFD) score as a metric to select the instruction data with high scores as diamond data.
% ({\romannumeral2})
\item
Then, we also introduce a novel Adversarial Instruction Output Embedding Consistency as the metric to mine the diamond data from online instruction data, via the adversairal prompt.
% ({\romannumeral3})
\item
Finally, we conduct an extensive set of experiments on two benchmark datasets to assess the performance. The experimental results serve to underscore the effectiveness of our proposed two methods. Moreover, the experimental results also show that it is of crucial practical significance to consider the robustness of prompts.
\end{itemize}

\section{Related Work}

\textbf{Instruction Data Mining.} While it's widely acknowledged that 'quality is paramount' in instruction tuning, the exploration of high-quality data beyond human curation remains largely uncharted. Instruction Mining~\cite{cao2023instruction} evaluates various metrics and employs a statistical regression model for data selection through the training of numerous models. In contrast, ALPAGASUS \cite{chen2023alpagasus} utilizes an external, fully-trained ChatGPT model to score each sample. While effective, this approach may overlook the intrinsic abilities of the base model, relying excessively on external models. Our work aims to develop a methodology that utilizes the representation feature of the target model to identify high-quality data for instruction tuning via the prompt's robustness, advancing the field with a more simple and efficient approach.

\textbf{Coreset Selection.}
The selection of a core set is pivotal in machine learning, aiming to identify a representative subset from massive datasets to expedite the learning process of various models. This approach has demonstrated its efficacy across various machine learning algorithms such as 
adversarial contrastive learning (ACL) ~\cite{NEURIPS2023_ef4f2a02}, 
continual learning ~\cite{NEURIPS2023_a0251e49}, % updated
% Support Vector Machines (SVM) learning~\cite{tsang2005core}, K-means ~\cite{har2005smaller}, and logistic regression~\cite{munteanu2018coresets}. 
and semi-supervised Learning~\cite{NEURIPS2021_793bc52a}. 
Recent advancements in neural network training, exemplified by studies such as Toneva et al.~\cite{TonevaSCTBG19}, delve into the dynamics of data point utility during training. They reveal that infrequently forgotten points have minimal impact on the final model accuracy. Paul et al.~\cite{paul2021deep} further showcase that expected loss gradient norm scores, averaged over various weight initialization, effectively prune training data without significantly compromising accuracy. 
Mindermann et al.~\cite{liu2018late} leverage Bayesian probability theory to estimate the individual impact of training points on holdout loss, refining training efficiency.
Wan et al.~\cite{liu2019multiple} extend the final similarity with the difference between dimensions to enhance the effectiveness of core set.
Additionally, Xia et al.~\cite{XiaL0S0L23} further consider the robustness of coreset within the change of different scenarios.

\section{Methodology}
\label{methodology}

In this section, we present a comprehensive overview of our proposed framework for mining diamond data from online instruction data, taking into account the influence of the robustness of online users' prompts on the quality of instruction data. We first introduce the background and notation for Instruction Following Difficulty in Section~\ref{Background}. Then we propose our AIFD for estimating online instruction data quality, via the instruction and response pairs in Section~\ref{AIFD}. Apart from it, we propose AIOEC for estimating online instruction data quality, through only the prompt of instruction data, considering the more challenge scene where the LLM's response frequently exhibit substantial disparities from human expectations in Section~\ref{AIOEC}.

\subsection{Background and notation for Instruction Following Difficulty}

\label{Background}

During the instruction-tuning phased, the loss of a sample pair $(Q,A)$ is determined through the continuous prediction of subsequent tokens, considering the instruction $Q$ and their proceeding words in the sequence:

\begin{equation}
L_{\theta}(A|Q)=-\frac{1}{N}\sum_{i=1}^{N}log P(w_{i}^{A}|Q, w_{1}^{A},w_{2}^{A},\cdots, w_{i-1}^{A} ;\theta)
\end{equation}
where $N$ is the number of words of the ground-truth answer $A$ and instruction $Q$ consists of both the prompt and the input. In addition, this averaged cross-entropy loss is denoted as the conditioned answer score $s_{\theta}(A|Q)=L_{\theta}(A|Q)$. This metric assessed the model's ability to generate suitable response according to given instruction $Q$.

However, the $s_{\theta}(A|Q)$ fails to consider the inherent difficulty of aligning $A$ itself. Therefore, in order to estimate the difficulty of following given sample instructions, a direct answer score $s_{\theta}(A)$ to measure the LLM's ability to generate the answer independently is introduced as following:
\begin{equation}
s_{\theta}(A) = -\frac{1}{N}\sum_{i=1}^{N} log P(w_{1}^{A},w_{2}^{A},\cdots, w_{i-1}^{A} ;\theta)
\end{equation}
And this score quantifies the inherent difficulty 
associated with autonomously generating an answer without explicit instructions. A higher score in the direct answer metric signifies a heightened level of difficulty in answer generation by the model.

To obtain better instruction data, identifying which instructions have a greater impact on the model while excluding the influence of the answers themselves, the concept of Instruction-Following Difficulty (IFD) score is proposed as follows:

\begin{equation}
r_{\theta}(Q,A)= \frac{s_{\theta}(A|Q)}{s_{\theta}(A)}
\end{equation}
Utilizing the IFD metric for data filtering has mitigated the impact of large models on fitting the answers themselves, allowing for a direct assessment of the influence of given instructions on the model's answer generation. Higher IFD scores indicate the model's inability to align the answer with the provided instructions, highlighting greater difficulty in instruction comprehension and presenting an advantageous opportunity for model refinement.

\subsection{Adversarial Instruction Following Difficulty score} 
\label{AIFD}
In this study, we delve into the impact of prompt's robustness within online instruction data, with a focus on instruction data mining. And we perform three different types of attacks: the character-level, word-level, and sentence-level attacks to generate adversarial prompts. In addition, the different types of attach processes are as follows:

% Prompts exhibiting strong robustness display minimal degradation in the quality of generated content when confronted with attacks, whereas prompts characterized by weak robustness significantly diminish the quality of generated content.

\textbf{Character-level:} We utilize the TextBugger~\cite{LiJDLW19} and DeepWordBug~\cite{gao2018black}, both of which manipulate texts by introducing typos or errors into words. This manipulation can include adding, deleting, repeating, replacing, and permuting characters within certain words.

\textbf{Word-level:} We employ the BertAttack~\cite{LiMGXQ20} and TextFooler~\cite{JinJZS20}, sophisticated techniques designed to replace words with synonyms or contextually similar alternatives, with the goal of deceiving large language models.

\textbf{Sentence-level:} We deploy the StressTest~\cite{NaikRSRN18} and CheckList~\cite{RibeiroWG021} methodologies, both of which involve appending irrelevant or extraneous sentences to prompts, aiming to distract LLMs. For example, in the CheckList attack, we generate 50 random sequences of alphabets and digits. Table \ref{tab:example} depicts the example of adversarial prompts generated by 6 attacks.

\begin{table}[ht]
\centering
\renewcommand\arraystretch{1.0}
\caption{
Example of adversarial prompts generated by 6 attacks.
}
\label{tab:example}
%\scriptsize
\begin{tabular}{ll}
\toprule\toprule
Clean & Give three tips for staying healthy. \\
\hline
TextBugger & Give three tips for staying \textcolor{red}{helthy}.  \\
\hline
DeepWordBug & \textcolor{red}{give} \textcolor{red}{threе} tips for \textcolor{red}{staуing} healthy.  \\
\hline
TextFooler & Give three tips for staying \textcolor{red}{salubrious.}\\
\hline
BertAttack & Give three \textcolor{red}{counseling} for staying healthy. \\
\hline
CheckList &  Give three tips for staying healthy \textcolor{red}{  zq0DcZ5dnI}.  \\
\hline
StressTest &  Give three tips for staying healthy \textcolor{red}{  and false is not true}.\\
\bottomrule\bottomrule
\end{tabular}
\end{table}

Our core motivation is to minimize cross-entropy loss during the model inference process, via the adversarial instruction data. In this paper, we propose the Adversarial Instruction Following Difficulty score to select high-quality data, a metric devised to evaluate the difficulty each adversarial instruction data presents.

Specially, we try to estimate the \textbf{Adversarial Instruction-Following Difficulty (AIFD) } scores $r_{\theta}(Q,A)$ on following adversarial instruction of six different given $(Q_{A},A) $pairs by calculating the ratio between $s_{\theta}(A)$ and $s_{\theta}(A|Q_{A})$:

\begin{equation}
r_{\theta}(Q,A, Q_{A})= \frac{s_{\theta}(A|Q)}{s_{\theta}(A)} + \sum_{i=1}^{i=6}\frac{s_{\theta}(A|Q_{A}^{i})}{s_{\theta}(A)}  ,Q_{A}^{i} \in Q_{A}
\end{equation}

By leveraging this score, the influence of LLM's intrinsic ability to fit the answer string is partially alleviated. The score measures the degree to which given adversarial instruction benefits the alignment of corresponding response. High AIFD scores infer the inability of the model to align response to the given corresponding adversarial instructions. which in turn indicates the difficulty of an adversarial instruction. Algorithm 1 describes the diamond data mining process from online instruction data via the AIFD approach.

\begin{algorithm*}[ht]
\label{algorithm}
\caption{Diamond data mining from online instruction data via the AIFD approach.}
  %\label{alg:Framwork}
    \begin{algorithmic}[1]
        \Require
        Online user's question list $\{Q_{1},Q_{2},\dots,Q_{N}\}$ , Total data number $N$, LLM $\mathcal{F}_{LLM}$  ;     
        \Ensure Diamond instruction data $D_{A_n}$;
        \For {$n$ in $N$ }
        %\label{code:fram:extract}
            % \State{ Input the user's question $Q_{n}$ into LLM $f$ to generate the corresponding response $A_{n}$ }
            % \State{ Construct online instruction data $D_{n}$ containing question $Q_{n}$ and the response $A_{n}$ }
            % \State{ Attack the prompt $P_{n}$ in the user's question $Q_{n}$ to generate adversarial prompt ${P_{A}}_{n}$ }
            % \State{ Construct adversarial instruction data ${D_{A}}_{n}$ containing${P_{A}}_{n}$ , $I_{n}$ and $A_{n}$ }
            % \State{ Calculate the AIFD score for the online instruction data $D_{n}$ according to the formula 4 }
    
            \State{$A_n = \mathcal{F}_{LLM}(Q_n)$ \hspace{2.2cm} // Generate corresponding response}
            \State{$D_n = \textbf{Construct}(Q_n, A_n)$ \hspace{1.0cm} // Construct online instruction data}
            \State{$P_{A_n} = \textbf{Attack}(P_n, Q_n)$ \hspace{1.35cm} // Attack the prompt to generate adversarial prompt}
            \State{$D_{A_n} = \textbf{Construct}(P_{A_n}, I_n, A_n)$ \hspace{0.1cm} // Construct adversarial instruction data}
            \State{$S_{n} = \textbf{Formula}_4(D_n)$ \hspace{1.65cm} // Calculate the AIFD score}
        \EndFor
        %\label{code:fram:trainbase}
    
        \State{$\{D_n\}=\textbf{Descend}(\{D_n\})$ \hspace{1.65cm} // Arrange online instruction data via AIFD scores}
        \State{$D_{core}=\textbf{Select}(\{D_n\})$ \hspace{2.0cm} // Select a certain proportion of $\{D_n\}$ as core data}   
        % \State{ Select a certain proportion of the online instruction data as the core data according to the sorted results from highest to lowest. }
        \State{$D^\prime_{core}=\textbf{Calibrate}(D_{core})$ \hspace{1.4cm} // Calibrate response to align with human preferences}
        % \State{ Expert calibration the response in the core data in order to align with human preferences}
        % \State{ Consider the calibrated core data as Diamond instruction data}
        \State{$D_{A_n}=\mathcal{G}(D^\prime_{core})$}
        % \State \Return Diamond instruction data 
        \State \Return $D_{A_n}$ 
    \end{algorithmic}
\end{algorithm*}

\subsection{Adversarial Instruction Output Embedding Consistency}
\label{AIOEC}

The AIFD method evaluates the quality of instruction data through the instruction and response pairs. However, during the inference process, LLMs frequently exhibit considerable disparities between model's response and human expectations. Due to the unreliable response, the AIFD method is no longer applicable for online data assessment. Consequently, we propose a novel AIOEC method to extract the high-quality instruction data only utilizing the prompt of instruction data.

\begin{figure*}[ht]
\centering
\includegraphics[scale=0.40]{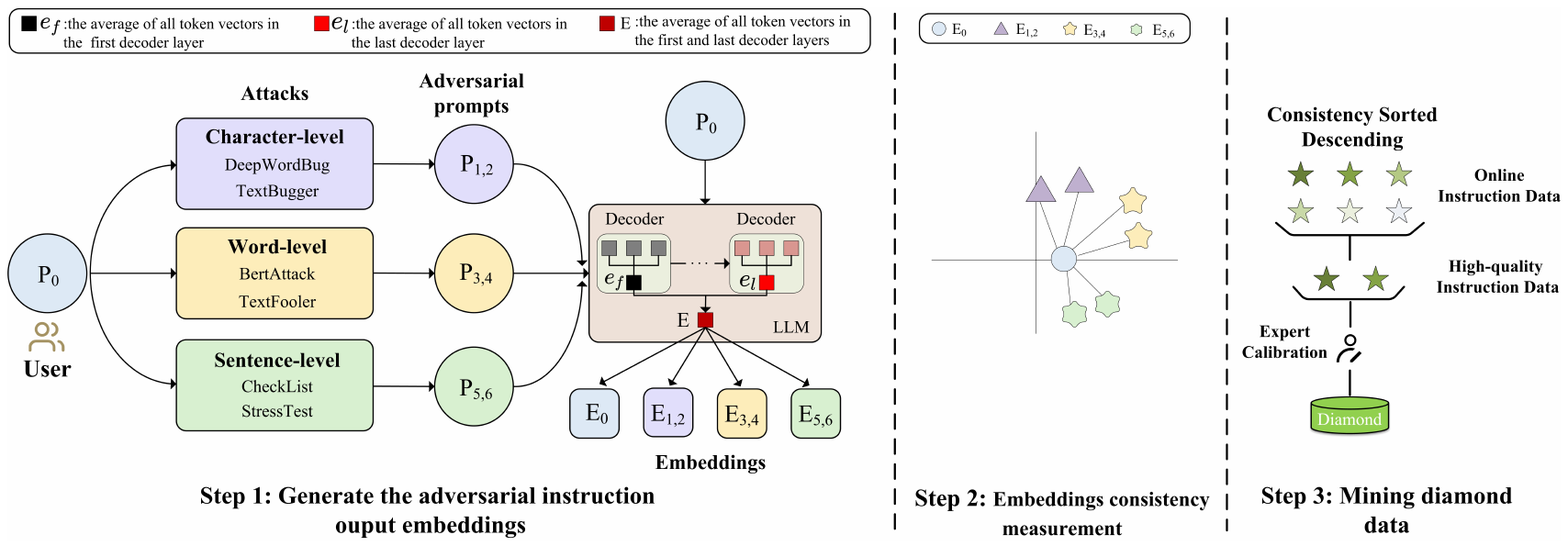}
\caption{ The illustration of Adversarial Instruction Output Embedding Consistency method for diamond data mining. And the method consists of three steps: generating adversarial instruction output embeddings, embeddings consistency measurement, and mining diamond data.}
\label{fig:illustration}
\end{figure*}

Figure \ref{fig:illustration} presents the illustration of Adversarial Instruction Output Embedding Consistency method for diamond data mining.  The AIOEC method consists of three steps. The first step is that we conduct the character-lever, word-level, and sentence-level attacks to generate six adversarial prompts for the prompt entered by online user.
Subsequently, inputting the six adversarial prompts as well as the prompt entered by the user into the LLMs to generate corresponding output embedding.

The second step entails assessing the consistency between the embeddings $E_{A}$ of outputs stemming from adversarial prompts and those originating $E$ from user-input prompt. The $E$ and $E_{A}$ are both derived from the summation of $e_{f}$ and $e_{l}$. And $e_{f}$ represents the average of all token vectors in the first layer of the LLMs, and $e_{l}$ represents the average of all token vectors in the last layer of the LLMs.

The consistency of adversarial prompt output embedding is calculated as following:

\begin{equation}
dis (E_{0},E_{A})= \sum_{i=1}^{i=6}cos(E_{0},E_{A}^{i}), E_{A}^{i} \in E_{A} 
\end{equation}

Finally, the quality of online instruction data is assessed based on this measure of consistency. Subsequently, the instruction data is ranked in descending order according to the metric scores. Ultimately, a certain proportion of the data is selected as "diamond data" for fine-tuning instructions.

\section{Experiments}
\label{experiments}

\subsection{Experimental Setting}

\textbf{Foundation Large Language Model.} To demonstrate the effectiveness of our proposed framework, we test it over two open-source large language models: LLaMA-7B ~\cite{touvron2023llama}, LLaMA2-7B~\cite{touvron2023llama2}.

\textbf{Evaluation and Datasets.} To assess the performance of the model for instruction tuning, we follow LLaMa's evaluation to perform zero-shot task classification on common sense reasoning datasets: MMLU~\cite{HendrycksBBZMSS21}, HellaSwag~\cite{zellers2019hellaswag}, ARC~\cite{clark2018think}, TruthfulQA~\cite{lin2021truthfulqa} on the OpenCompass assessment framework~\cite{2023opencompass}.

\textbf{Training Datasets} The Alpaca datasets~\cite{taori2023stanford} encompass 52002 instruction-following samples. WizardLM dataset~\cite{xu2023wizardlm} leverages the EvoInstruction data. The incorporation of ChatGPT during the reformulation guarantees high fidelity of data. We utilize the WizardLM70K for our experiment.

\subsection{Implementation Details}
\label{sec_im}
In our experiments with the pre-trained LLaMA-7B and LLaMA2-7B models fine-tuned utilizing the diamond data, we adopted a training configuration consistent with that of the original Alpaca and WizardLM, leveraging the llm-action codebase. And the detailed training specifications can be found in Appendix ~\ref{sec_ap_im}.

\subsection{Experimental Results}
\label{sec_er}

\subsubsection{The Diamond Data Mining from Offline Instruction Datasets for Instruction Tuning}

To demonstrate the effectiveness of our approach, which emphasizes the significance of considering prompt robustness in facilitating diamond data mining, we firstly conducted experiments on the offline instruction datasets.

\begin{table}[ht]
% \scriptsize
\renewcommand\arraystretch{1.2}
\caption{The comparison of performance for fine-tuned LLaMA-7B and LLaMA2-7B incorporating Alpaca data of different proportions as diamond data on different tasks.}
\label{tab:alpaca}
\centering
\resizebox{\linewidth}{!}{
    \begin{tabular}{ c | c | c c c c c  | c c c c  c }
    \toprule\toprule
    \multirow{2}{*}{Datasets}  & \multirow{2}{*}{Methods}   &  \multicolumn{5}{c|}{LLaMA-7B} &  \multicolumn{5}{c}{LLaMA2-7B}\\
    & & ARC & HellaSwag & MMLU & TruthfulQA & Average &  ARC & HellaSwag & MMLU & TruthfulQA & Average \\
    \midrule
    &Pre-train  &34.27 & 74.31 & 35.53 & 34.31 &  44.61 & - & -& - & -& -  \\
    \midrule
    \multirow{2}{*}{5\% Alpaca}&IFD  & 36.80& 76.60 & 37.05 & 36.92 & 46.84    &  38.29& 76.65 & 46.00 & 47.11 &   52.01  \\
    &AIFD  & \textbf{38.43} & \textbf{76.72} & \textbf{37.54} & \textbf{39.45} & \textbf{48.04}& \textbf{39.17} & \textbf{76.89} & \textbf{46.65} &  \textbf{47.47}& \textbf{52.55}     \\
    \midrule
    \multirow{2}{*}{10\% Alpaca}&IFD  & 37.57& 76.56 & 36.75 & 37.41 & 47.07 &  37.48 & 76.47 & 45.45 & 46.70 & 51.53\\
    &AIFD  & \textbf{40.10} & \textbf{76.64} & \textbf{37.96} & \textbf{38.73} & \textbf{48.36} &\textbf{40.64} & \textbf{76.60} & \textbf{45.55}& \textbf{47.28} & \textbf{52.52}\\
    \midrule
    \multirow{2}{*}{15\% Alpaca}&IFD  & 37.40& 77.45 & 35.18 & 38.05 & 47.02  &  43.74 & 76.40 & 47.18 &48.02 & 53.84\\
    &AIFD  & \textbf{42.18} & \textbf{78.49} & \textbf{36.22} & \textbf{39.23} & \textbf{49.03} &\textbf{48.71} & \textbf{76.72} & \textbf{47.54} &\textbf{48.61}  & \textbf{55.40}\\
    \midrule
    \multirow{2}{*}{20\% Alpaca}&IFD  & 38.38& 79.24 & 38.60 & 39.69 & 48.98 &44.93 & 76.47& 47.22&   46.56 &  53.80 \\
    &AIFD  & \textbf{38.78} & \textbf{79.46} & \textbf{40.42} & \textbf{41.05} & \textbf{49.93}& \textbf{45.67} & \textbf{76.60} & \textbf{47.41} &  \textbf{46.50}& \textbf{54.00}\\
    \midrule
    
    Official Alpaca& &42.48& 76.91& 42.16&39.55 &50.25 & 50.23 & 77.65 & 46.75 & 44.87&54.88\\
    \bottomrule\bottomrule

    \end{tabular}
}

\end{table}

Table \ref{tab:alpaca} illustrates the comparison of performance for fine-tuned LLaMA-7B and LLaMA2-7B incorporating Alpaca data as diamond data across for different tasks, via the opencompass evaluation tool. According to Table \ref{tab:alpaca}, we can observe that our proposed AIFD approach achieves \textbf{1.2\%} higher average accuracy than IFD approach on 4 different tasks, only utilizing 5\% diamond data to fine-tune the LlaMA-7B model. In addition, after fine-tuning with diamond data mined via our AIFD approach, the LLaMA-7B' performance increased by \textbf{4.43\%} compared to the pre-trained LLaMA-7B model, slightly lower than the LLaMA-7B with official Alpaca, which was \textbf{2.21\%}.

Furthermore, we curate subsets comprising the top 10\%, 15\% and 20 \% of the Alpaca datasets as the diamond data to fine-tune the LLaMA-7B model, facilitating an investigation into performance fluctuations. As depicted in Table \ref{tab:alpaca}, we present the performance comparison of LLaMA-7B fine-tuned at various proportions of Alpaca datasets. A consistent finding emerges: LLaMA-7B consistently outperforms the IFD method when fine-tuning with data from the Alpaca datasets using our proposed AIFD approach. In addition, the experimental results show that we can select more high-quality instruction data for instruction tuning, via our AIFD method, considering the impact of prompt's robustness on instruction data.

To further verify the superior performance of our proposed method for high-quality instruction mining, we conducted experiments for LLaMa2-7B by leveraging the Alpaca datasets. According to the Table \ref{tab:alpaca}, the experimental results show that the performance improvement of the LLaMA2-7B fine-tuned with 5\% diamond data based on our AIFD method compared to IFD is \textbf{0.54\%}.

\begin{wraptable}{r}{9.5cm}
\vspace{-4.5mm}
\renewcommand\arraystretch{1.2}
% \scriptsize
\caption{The comparison of performance of LLaMa-7B fine-tuned with WizardLM70K Data at various ratios for different tasks.}
\label{tab:wizard}
\centering
\vspace{1.5mm}
\resizebox{\linewidth}{!}{
    \begin{tabular}{ c | c | c c c c| c}
    \toprule\toprule
    
    Datasets& Methods & ARC & HellaSwag & MMLU & TruthfulQA & Average \\
    \midrule
    
    \multirow{2}{*}{10\% WizardLM70K}  &IFD  &37.56& 75.35 & 36.25 & 43.89 & 48.26 \\
    &AIFD  & \textbf{41.62} & \textbf{75.69} & \textbf{37.18} & \textbf{45.23}  & \textbf{49.93}\\
    \midrule
    \multirow{2}{*}{20\% WizardLM70K} &IFD& 43.58 & 74.82 & 37.78 & 44.63 & 50.20\\
    &AIFD & \textbf{44.82} & \textbf{74.93} & \textbf{38.57} & \textbf{44.62} & \textbf{50.74} \\
    \midrule
    \multirow{2}{*}{40\% WizardLM70K} &IFD  &  46.03 &73.38 & 39.57 & 45.04 &51.00 \\
    &AIFD & \textbf{47.62} & \textbf{73.55} & \textbf{40.14} &  \textbf{45.65}& \textbf{51.74} \\
    \bottomrule\bottomrule
    
    \end{tabular}
}
\vspace{-2mm}
\end{wraptable}

Moreover, we also conducted instruction tuning experiments on the WizardLM70K dataset. We craft subsets containing the top 10\%, 20\% and 40 \% of the WizardLM70K datasets as the diamond data to fine-tune the LLaMA-7B model, enabling us to investigate the performance changes.
Table \ref{tab:wizard} depicts the comparison of performance of LLaMa-7B fine-tuned with WizardLM70K Data at various ratios across different tasks. According to Table \ref{tab:wizard}, we can observe that our proposed AIFD approach achieves \textbf{1.67\%} higher average accuracy than IFD approach on 4 different tasks, only utilizing 10\% diamond data to fine-tune the LlaMA-7B model.

\subsubsection{The Diamond Data Mining from Online Instruction Datasets for Instruction Tuning}

\begin{table}[ht]
\renewcommand\arraystretch{1.2}
% \scriptsize
\caption{The comparison of performance for fine-tuned LLaMA-7B and LLaMA2-7B incorporating online Alpaca data of different proportions as diamond data on different tasks.}
\label{tab:alpacaonline}
\centering
\resizebox{\linewidth}{!}{
    \begin{tabular}{c | c | c c c c c | c c c c c}
    \toprule\toprule
    \multirow{2}{*}{Datasets}  & \multirow{2}{*}{Methods}   &  \multicolumn{5}{c|}{LLaMA-7B} &  \multicolumn{5}{c}{LLaMA2-7B}\\
    & & ARC & HellaSwag & MMLU & TruthfulQA & Average &  ARC & HellaSwag & MMLU & TruthfulQA & Average \\
    \midrule
    
    \multirow{3}{*}{5\% Alpaca} &IFD  & 36.54& 76.08 & 37.02 & 38.08& 46.93  &40.64&  74.79 &  44.27 & 45.80&51.38\\
    &AIFD   &\textbf{39.76}&\textbf{   75.82}&\textbf{  36.68 }&\textbf{ 38.62 }&\textbf{ 47.72 }&\textbf{ 50.29  }&\textbf{ 73.99 }&\textbf{ 46.85 }&\textbf{ 43.64 }&\textbf{ 53.69}\\
    & AIOEC  &\textbf{ 37.36 }&\textbf{ 76.74 }&\textbf{ 36.96 }&\textbf{  38.37}&\textbf{ 47.63}&\textbf{42.32 }&\textbf{ 74.65 }&\textbf{ 46.55 }&\textbf{44.03  }&\textbf{ 51.89}\\
    \midrule
    
    \multirow{3}{*}{10\%Alpaca }& IFD  & 38.09& 76.31 & 37.69 & 38.35& 47.61 & 51.88&  74.76& 47.25 &44.63&  54.63\\
    &AIFD &\textbf{ 38.58}&\textbf{  76.57}&\textbf{  38.26}&\textbf{ 39.16}&\textbf{  48.14 }&\textbf{51.88}&\textbf{  74.76}&\textbf{ 47.25 }&\textbf{44.63}&\textbf{  54.63} \\
    & AIOEC  &\textbf{ 40.93 }&\textbf{ 76.46 }&\textbf{ 36.52 }&\textbf{ 38.08 }&\textbf{  48.00 }&\textbf{49.75 }&\textbf{ 74.82 }&\textbf{ 47.44 }&\textbf{ 44.24 }&\textbf{ 54.06}\\
    \midrule
    \multirow{3}{*}{15\% Alpaca}&IFD  &40.23 & 74.78 &  38.14 & 39.78& 48.23 & 42.33  & 75.54 & 46.14  & 46.44& 52.61\\
    &AIFD  &\textbf{ 44.43 }&\textbf{ 75.67 }&\textbf{ 38.68 }&\textbf{ 40.55}&\textbf{  49.83 }&\textbf{53.31 }&\textbf{ 75.19}&\textbf{ 48.11 }&\textbf{ 44.81 }&\textbf{  55.36} \\
    & AIOEC  &\textbf{  45.00}&\textbf{ 76.21 }&\textbf{ 37.18 }&\textbf{38.90  }&\textbf{ 49.32}&\textbf{ 55.55 }&\textbf{  75.07}&\textbf{ 47.32 }&\textbf{44.44  }&\textbf{ 55.60}\\
    \midrule
    \multirow{3}{*}{20\% Alpaca}&IFD  &43.80& 75.02 &  37.85 & 40.67&  49.34  &42.36& 75.33 &  45.44 & 45.13& 50.07  \\
    &AIFD  &\textbf{ 45.01}&\textbf{ 75.78 }&\textbf{ 39.45 }&\textbf{  40.16 }&\textbf{  50.10 }&\textbf{ 57.31}&\textbf{ 75.45 }&\textbf{47.78 }&\textbf{  44.90  }&\textbf{  56.36}\\
    & AIOEC  &\textbf{46.42}&\textbf{ 75.95 }&\textbf{38.21  }&\textbf{39.64   }&\textbf{  50.01 }&\textbf{ 53.96}&\textbf{ 74.98 }&\textbf{  47.21}&\textbf{  41.69 }&\textbf{  54.46}\\
    \bottomrule\bottomrule
    \end{tabular}
}
\end{table}

We delve further into exploring the efficacy of our proposed methods in more challenging scenario, specifically in selecting high-quality data from online instruction data. In addition, the comprehensive procedure for generating online command data can be located in Appendix~\ref{sec_ap_cd}. We utilized AIFD and AIOEC methods to mine high-quality from online instruction data. Subsequently, we replaced the responses in this high-quality data with the real answers corresponding to the instructions from the offline data, for the purpose of instruction tuning.

Table \ref{tab:alpacaonline} compares the performance for fine-tuned LLaMA-7B and LLaMA2-7B incorporating online Alpaca data of different proportions as diamond data on different tasks.  Using our proposed methods, we selected 5\% of the online data as diamond data for fine-tuning LLaMA-7B and LLaMA2-7B. Both models show superior performance compared to the baseline IFD method.

\iftrue{
\begin{figure*}[htbp]
    \centering
    \subfloat[10\% WizardLM70K]{\includegraphics[width=0.42\textwidth]{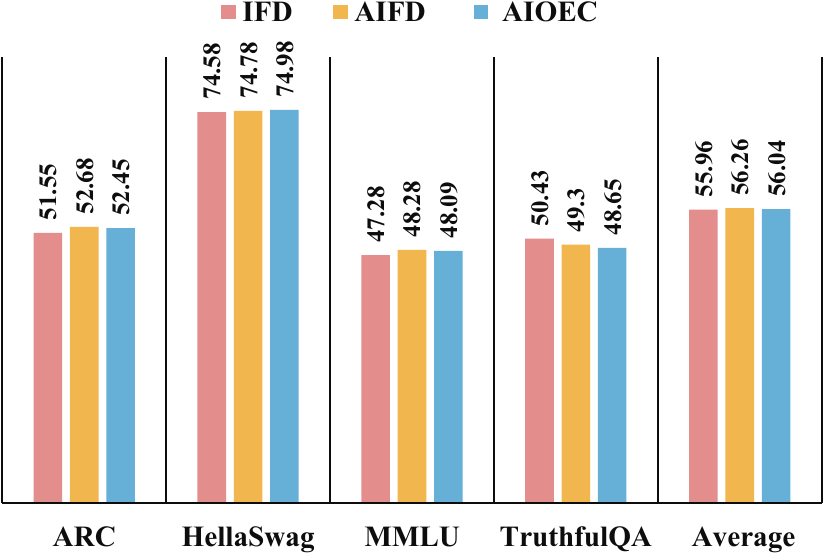}}
    \hfill
    \subfloat[20\% WizardLM70K]{\includegraphics[width=0.42\textwidth]{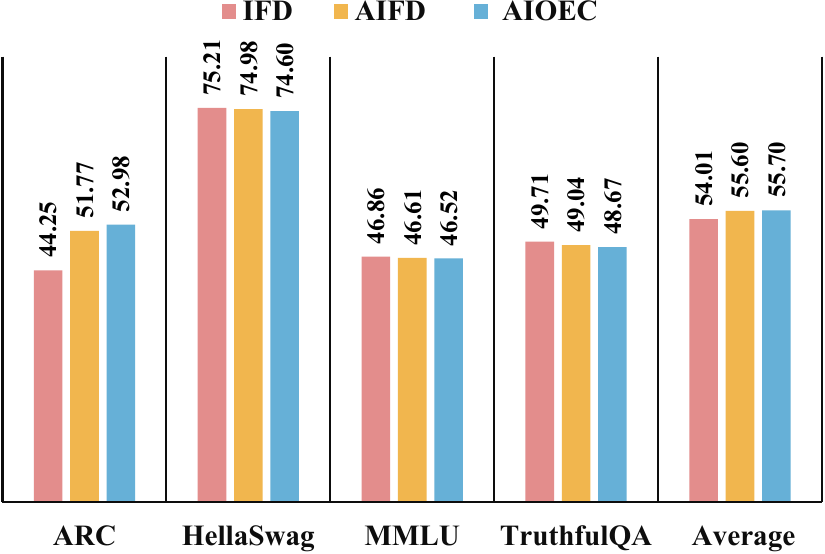}}
    \caption{The comparison of performance for the fine-tuned LLaMa2-7B utilizing 10\% and 20\% WizarrdLM70K Datasets on the different tasks.}
    \label{fig:wizard}
\end{figure*}
}\fi

Figure \ref{fig:wizard} presents the comparison of performance for the fine-tuned LLaMa2-7B utilizing 10\% and 20\% WizarrdLM70K Datasets on the different tasks. And the results demonstrate that our proposed methods can mine more high-quality instruction data.

\subsection{Ablation study}

\begin{wraptable}{r}{8cm}
    \vspace{-4.5mm}
    \renewcommand\arraystretch{1.2}
    % \scriptsize
    \caption{The comparison of performance for the fine-tuned LLaMa2-7B utilizing 5\% Alpaca data on the different tasks by different attacks.}
    \label{tab:5alpaca}
    \centering
    \resizebox{\linewidth}{!}{
        \begin{tabular}{  c  | c  c c c c}
        \toprule\toprule
          
         & ARC & HellaSwag & MMLU & TruthfulQA & Average \\
        \midrule
        
        Three attacks & 43.89 &  75.16&  46.52 & 45.30& 52.72\\
        Character attack & 39.95 & 72.31 &  44.84 & 39.09&  49.05\\
        Word attack  &39.91 & 75.77 &46.14 & 45.01 & 51.71\\
        Sentence attack & 38.83 & 72.84& 45.19 &39.77 &  49.16\\
        
        \midrule
        
        IFD + Character  &39.17  &76.74 & 46.67 &  47.47 & 52.51 \\

        IFD + Word  &39.17  &76.74 & 46.67 &  47.47 & 52.51     \\
        
        IFD + Sentence  &39.17 & 76.74 &  46.67& 47.47  &  52.51    \\
        \bottomrule\bottomrule
        
        \end{tabular}
    }
\end{wraptable}

To further verify the impact of prompt robustness for instruction data mining, we conducted additional training and evaluation on our proposed diamonds data framework. 
Building upon the baseline IFD method, we incorporate various types of attacks and then select the top 5\% of instruction data as the diamond data for fine-tuning LLaMA2-7B, via the AIFD score. The performance evaluation results of the fine-tuned LLaMA2-7B are presented in Table \ref{tab:5alpaca}. According to the experimental results in Table \ref{tab:5alpaca}, it can be observed that incorporating various attacks on the foundational IFD method yields consistent model performance. This indicates that introducing various types of attacks yields the same diamond data when filtered using the AIFD method. It also suggests that different types of attacks have an equivalent effect on data selection. More importantly, we found that fine-tuning LLaMA2-7B with instruction data mined from just three types of attacks resulted in the best model performance.

\subsection{Diamond Data Characteristics}

\begin{wrapfigure}{r}{6.4cm}
    \vspace{-9.6mm}
    \centering
    \resizebox{\linewidth}{!}{
    \includegraphics[scale=0.8]{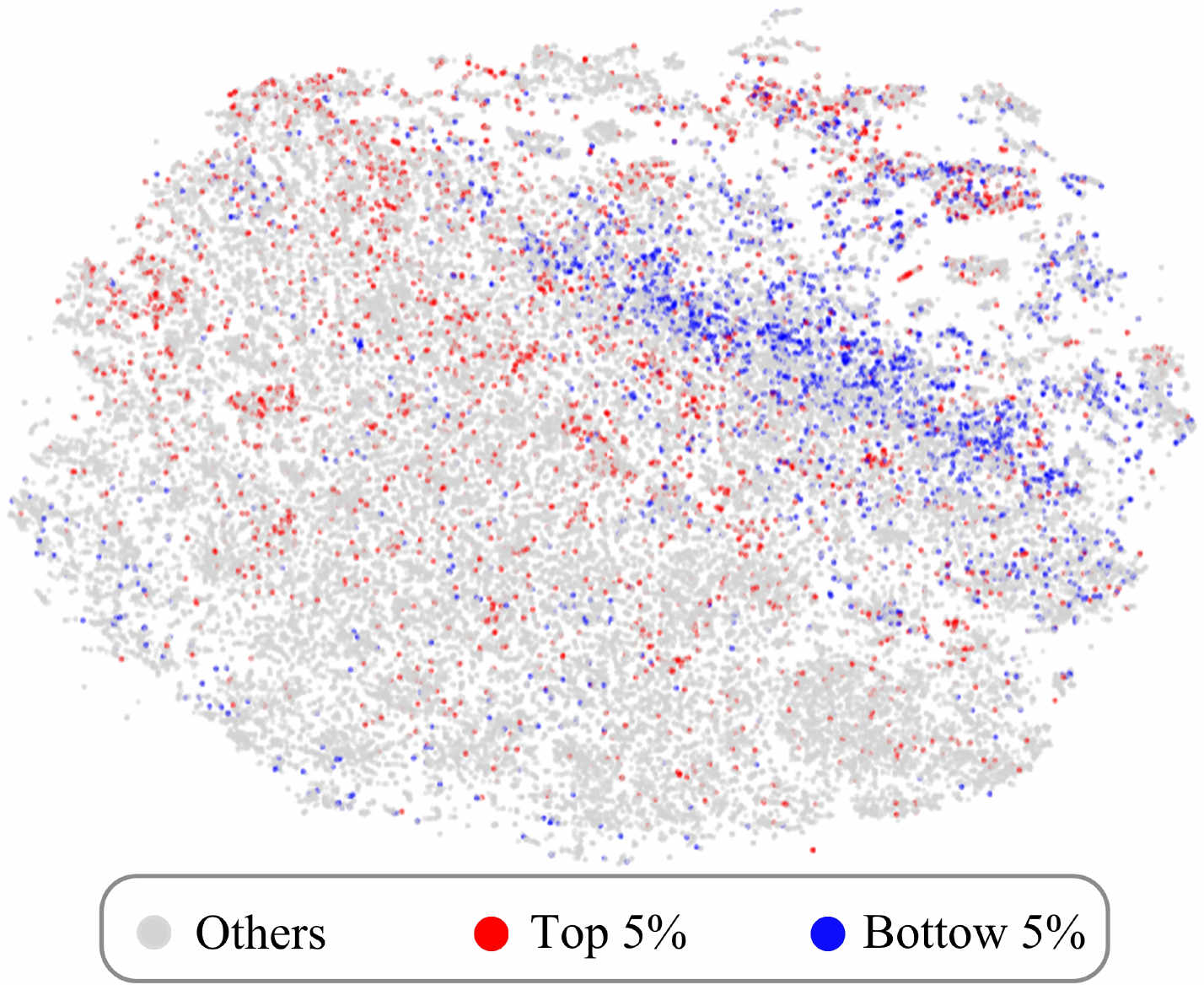}
    }
    \caption{Visualization using t-SNE on instruction embeddings from the Alpaca dataset. }
    \label{fig:state}
    \vspace{-5mm}
\end{wrapfigure}

\textbf{Distribution Characteristics.} In this section, our primary focus lies in comprehensively understanding the distributional characteristics of the diamond data within the original dataset. Initially, we calculate the embedding of each instruction in the Alpaca dataset using the pre-trained LLaMA-7B model. Subsequently, we employ t-SNE~\cite{van2008visualizing} for dimensionality reduction, thereby mapping high-dimensional embeddings to a 2D space. The visualized vectors, color-coded according to the top or bottom 5\% difficulty ratios, are displayed in Figure \ref{fig:state}. Contrary to conventional beliefs, our carefully selected instruction exhibits non-uniform dispersion. Instead, discernible boundaries emerge between samples of high and low difficulty, challenging prior assumptions that selected data should span the entire spectrum of instructions and maximize diversity.

\textbf{Pattern Characteristics.} To delve deeper into the pattern characteristics of the diamond data under scrutiny, we utilized the Berkeley Neural Parser~\cite{kitaev2018constituency}. This sophisticated tool  enables us to effectively delineate the verb-noun structure inherent in the instructions accompanying each data sample. Through this analytical framework, we are able to pinpoint the principal verb along with its corresponding direct noun object in each instruction. Consequently, this approach affords us direct insights into the types of instructions predisposed to receiving higher or lower AIFD scores. Our experimental methodology draws upon the Alpaca dataset, wherein we examine the top 10 verb-noun pairs derived from the highest and lowest 5\% AIFD scores, as presented in Table \ref{tab:verbnoun}.

\begin{wraptable}{r}{9cm}
\vspace{-3.8mm}
% \scriptsize
\renewcommand\arraystretch{1.2}
\caption{The top 10 occurred verb-noun pairs from the
 top5\% AIFD scores data and the low5\% AIFD scores data.}
\label{tab:verbnoun}
\centering
\vspace{1mm}
\resizebox{\linewidth}{!}{
    \begin{tabular}{  c c c | c c c }
    \toprule\toprule
      \multicolumn{3}{c|}{  Top 5\% AIFD}    &\multicolumn{3}{c}{ Low 5\%AIFD  }  \\
      Verb& Noun & Count & Verb& Noun & Count  \\
    \midrule
    Write& story&35 &make & sentence & 88  \\
     Generate & story &24 &Rewrite&sentence & 78  \\
     
    Write & essay&22 & Edit&sentence &56   \\
    
    Generate& list&14 & Change&sentence &29  \\
    Write & article&13 & be& sentence&29   \\
     Create& story&12 & use& sentence& 25  \\
     
     Write&poem &12 &make &text &20   \\
     Write& importance &11 & go& sentence &19   \\
     Write& post &11 & Take& sentence &19\\
     Generate& word&10 & correct& sentence &18 \\
    \bottomrule\bottomrule
    \end{tabular}
}
\vspace{-6mm}
\end{wraptable}

From this experiment, a clear disparity emerges in the pattern characteristics between high-AIFD data and low-AIFD data. High-AIFD data primarily comprises creative and intricate instructions such as "write story," "generate story," and "write essay," necessitating substantial creativity, analytical skills, and profound comprehension. Conversely, low-AIFD data revolves more around adherence to guidelines and demands comparatively less creativity, showcasing a spectrum in the cognitive demands and creativity required by different tasks for language models. Consequently, the validity of AIFD as a metric for data filtration can be succinctly summarized by its efficacy in identifying instructions that entail heightened levels of creativity and profound understanding.

\subsection{Broader Impacts}
\label{sec_br}

This paper proposes related technologies that can effectively evaluate the robustness for instruction data, strengthen LLMs' application in various scenarios, and enhance its ability to cope with user misuse, thereby further improving the effectiveness of LLMs usage. To the best of our knowledge, the techniques proposed in this paper, do not lead to any negative societal impacts.

\section{Conclusion and limitation}
\label{con}
Our study unveils the potential of leveraging the robustness of prompt to mine high-quality instruction data that aligns with the model. In this study, we propose a novel framework designed for online instruction data mining, integrating AIFD and AIOEC approaches. We conducted extensive experiments on Alpaca and WizarrdLM-70k datasets, revealing the superior performance of the proposed approaches. Moreover, the results underscore the critical practical significance of considering prompt‘s robustness. 

To the best of our knowledge, this is the first investigation into the impact of prompt's robustness for online instruction mining. We only analyze the impact of prompt robustness on high-quality data selection through generating adversarial instruction data by attacking prompts. We aspire this paper to inspire fellow researchers to delve into the prompt's robustness for online instruction mining.

\clearpage

%\newpage
\bibliography{neurips_2024}
\bibliographystyle{unsrt}

\clearpage

\appendix

\section{Implementation Details}
\label{sec_ap_im}

For experiments conducted on the pre-trained LLaMA-7B model, our training framework adheres to protocols similar to those used in the Alpaca and WizardLM datasets. During the experiments, the model was trained using the Adam optimizer with a batch size of 128, spanning three epochs for training. When training on the Alpaca dataset, a maximum input length of 512 was employed. However, due to hardware limitations, we opted for a 1024 input length when using the WizardLM dataset, while the original model utilizes a length of 2048, providing it with an inherent advantage. In experiments involving the LLaMA2-7B model, we employed instruction prompts derived from Vicuna. And in this paper, all experiments are based on the eight V100 graphics cards and the PyTorch 2.1 framework.

\section{Online Instruction Data Construction Details}
\label{sec_ap_cd}

In our study, we utilized the instruction $Q$ from Alpaca and WizarrdLM70K datasets, along with LlaMA 7B and LlaMA2-7B models fine-tuned using a small amount of instruction data, to generate the online response. In addition, we constructed online instruction data by pairing instruction $Q$ with the online response. Table 7 presents the examples for online instruction data.

The small amount of instuction data is selected as following: we employ the basic clustering technique K-Means on the embeddings of these instructions. Specifically, we create 100 clusters based on instruction embeddings and randomly select 10 instances from each cluster. Subsequently, the initial model undergoes training for only one epoch using these samples to derive our preliminary pre-trained LlaMA 7B and LlaMA2-7B models.

\iftrue{
\begin{sidewaystable*}[ht]
\label{tab7}
\renewcommand\arraystretch{1.2}
\caption{Examples for online instruction data.}
\centering
\begin{tabular}{  c |c| c }
\hline  
& Instruction &Response \\
\hline
Offline Alpaca&  Give three tips for staying healthy.& 
\makecell{
1.Eat a balanced diet and make sure to\\ include plenty of fruits and vegetables. \\
2. Exercise regularly to keep \\your body active and strong. \\
3. Get enough sleep and maintain \\a consistent sleep schedule. }\\

\hline

\makecell{Online Alpaca\\ for LlaMA-7B}& Give three tips for staying healthy. &  \makecell{1. Get enough sleep. \\2. Exercise regularly. \\3. Eat a balanced diet.}\\

\hline

\makecell{Online Alpaca\\ for LlaMA2-7B}&  Give three tips for staying healthy. & \makecell{Three tips for staying healthy are: \\eat a balanced diet, get regular exercise,\\ and get enough sleep.}\\

\hline

\makecell{Online WizarrdLM70K\\ for LlaMA-7B} & 
\makecell{
Can you provide a list of healthy habits to maintain a healthy lifestyle? \\
Please format your response as an HTML page with bullet points.\\
<html><body><h3>Healthy Habits:</h3>\\    
<ul><li>Eating a balanced diet with plenty of fruits and vegetables.</li>\\      
<li>Engaging in regular physical activity, \\such as walking, running, or cycling.</li>\\ 
<li>Getting enough sleep each night, ideally 7-8 hours.</li>\\      
<li>Staying hydrated by drinking plenty of water throughout the day.</li>\\      
<li>Limiting alcohol consumption and avoiding smoking.</li>\\      
<li>Managing stress through relaxation techniques \\like meditation or yoga.</li>\\
<li>Regularly visiting a healthcare provider for check-ups 
\\and preventative care.</li></ul></body></html>ideally} &   
\makecell{
Here are some healthy habits to \\maintain a healthy lifestyle: \\
1. Eat a balanced diet with plenty of \\fruits, vegetables, and whole grains. \\
2. Get regular exercise, such as \\walking, running, or swimming. \\
3. Get enough sleep each night,  7-9 hours.\\ 
4. Limit your alcohol intake and avoid smoking. \\
5. Manage stress}\\
\hline

\makecell{Online WizarrdLM70K\\ for LlaMA2-7B}& 
\makecell{
Can you provide a list of healthy habits to maintain a healthy lifestyle? \\
Please format your response as an HTML page with bullet points.\\
<html>  <body>    <h3>Healthy Habits:</h3>\\    
<ul>      <li>Eating a balanced diet with plenty of fruits and vegetables.</li>\\      <li>Engaging in regular physical activity, \\
such as walking, running, or cycling.</li>\\      
<li>Getting enough sleep each night, ideally 7-8 hours.</li>\\      
<li>Staying hydrated by drinking plenty of water throughout the day.</li>\\      
<li>Limiting alcohol consumption and avoiding smoking.</li>\\      
<li>Managing stress through relaxation techniques \\like meditation or yoga.</li>\\      
<li>Regularly visiting a healthcare provider for check-ups and \\
preventative care.</li></ul></body></html>ideally }&   
\makecell{
To maintain a healthy lifestyle, \\consider adopting these habits:\\
Balanced Diet: Ensure your meals include \\
plenty of fruits, vegetables, and whole grains\\
to provide essential nutrients. \\
Regular Exercise: Incorporate activities like \\
walking, running, or swimming into your routine\\
to promote physical fitness. \\
Adequate Sleep: Aim for 7-9 hours of sleep each \\
night to allow your body to rest and rejuvenate. \\
Moderate Alcohol Consumption: Limit alcohol intake \\
and avoid smoking to protect your overall health. \\
Stress Management: Practice stress-reducing \\
techniques such as meditation or deep breathing \\
exercises to maintain mental well-being.
}\\

\hline

\end{tabular}
\label{tab:ins&res}
\end{sidewaystable*}
}\fi

\end{CJK}
\end{document}